# Social Media Sentiment Analysis for Cryptocurrency Market Prediction

Ali Raheman[1], Anton Kolonin[1,2,3][0000-0003-4180-2870], Igors Fridkins[2],
Ikram Ansari[1][0000-0002-9091-6674], Mukul Vishwas[1][0000-0002-8824-1954]

[1] Autonio Foundation Ltd., Bristol, UK
[2] SingularityNET Foundation, Amsterdam, Netherlands
[3] Novosibirsk State University, Novosibirsk, Russian Federation
`{ali.raheman,akolonin}@gmail.com`

**Abstract.** In this paper, we explore the usability of different natural language processing models for the sentiment analysis of social media applied to financial market prediction, using the cryptocurrency domain as a reference. We study how the different sentiment metrics are correlated with the price movements of Bitcoin. For this purpose, we explore different methods to calculate the sentiment metrics from a text finding most of them not very accurate for this prediction task. We find that one of the models outperforms more than 20 other public ones and makes it possible to fine-tune it efficiently given its interpretable nature. Thus we confirm that interpretable artificial intelligence and natural language processing methods might be more valuable practically than non-explainable and non-interpretable ones. In the end, we analyze potential causal connections between the different sentiment metrics and the price movements.

**Keywords:** Cryptocurrency, Explainable Artificial Intelligence, Financial Market, Interpretable Artificial Intelligence, Natural Language Processing, Sentiment Analysis

## 1      Introduction

We all are well aware of how much social media is connected to everyone's life and the impact it has on it. Recently we have witnessed how tweets/news can change the dynamic pricing of cryptocurrencies. With this in mind, we tried to determine how sentiment is correlated to the price change and whether it is possible to predict? In simple words, can a sentiment score be used for price prediction. The Natural Language Processing (NLP) domain has developed so fast that we have better ways to deal with raw texts than in recent years.

With easily accessible technologies, it is now very easy to put your thoughts out on social media in the form of blog posts, online forums, reviews, and feeds (such as Twitter or Reddit). This leads us to an astonishing number of texts every second; a recent study shows an average person produces 1.7 MB of data every second, 102 MB in a minute and we all send a total of 18.7 Billion texts every day. We focused on Twitter and Reddit for our text data sources and collected about six months of data for our experiments. In this paper, we compare different machine learning (ML) models as well as models based on lexicons and "n-grams" and analyze their performance. People are creating various communities to spread their thoughts and others follow a person/page to get more insight into a particular domain.

Social media provides diverse exposure to business and the various ways to connect to their customers. Consumers can use the product or service and can provide feedback (reviews) of said service/product. Sentiment analysis is widely used to extract valuable insights from the received feedback, which can help improve or evolve the service/product for future customers.

Twitter/Reddit are the types of social media where anyone can express their thoughts, reviews, memes or daily life events. These tweets and feeds can affect the cryptocurrency markets due to the large number of people who are deeply into the cryptocurrency markets and publish technical analyses and thoughts of the markets. Therefore, they become 'reference' sources of thoughts/analyses which leads to a majority of people following them. With this information, it is clear to say the feedback/thoughts from social media are very important and can help create a better-involving prediction of the price movements.

Sentiment analysis was first used in the 1950s and the field has been continuously evolving ever since. In this research, we have tried to evaluate more than twenty different sentiment analysis models found in the public domain and evaluated them with respect to cryptocurrency-specific text corpus based on the latest public Twitter and Reddit news feeds. We have found the superior model based on "n-grams" associated and were able to improve its performance significantly due to its "interpretable" nature, as such, we could amend and extend vocabularies of entries corresponding to positive and negative sentiment in custom cryptocurrency-specific jargon.

## 2    Methodology

This study has been divided into five parts. First, a literature survey was conducted on all publicly available sentiment models as identified further. Second, the six months worth of public Twitter and Reddit "tweets" and posts across 77 well-known feeds/subreddits in the cryptocurrency community have been collected. Third, the collected data was processed using each of the identified models and their comparative performances have been evaluated. Fourth, after completing the third phase, we found the best model for the financial domain was the open-source Aigents model (identified as "aigents" in Figure 1). Then, we improved the vocabularies of n-grams of the latter model and re-evaluated the performance of the models. At this point, we have found the correlation between improved Aigents model sentiment score and the "ground truth" significantly increased from 0.33 to 0.57 (identified as "aigents+" in Figure 1). Fifth and finally, we have explored possible causal connections between the sentiment metrics and the price movements studying mutual Pearson correlation between daily Bitcoin price difference (derivative) and each of the basic four sentiment metrics (*sentiment, positive, negative, contradictive*) discussed further, all metrics are aggregated on daily basis.

## 3    Sentiment Analysis Models

Sentiment Analysis (SA), also known as opinion analysis or emotion AI, can be defined as the process of calculating emotions, opinions, and attitudes scores. This score can be used for further analysis and usually the sentiment scores are 'Positive', 'Nega-tive', and 'Neutral'. Sentiment analysis problems may be further addressed from a few different perspectives, as follows.

### 3.1   Fine-grained sentiment

This is the most simplified sentiment analysis task to understand, consisting mostly of the customer's feedback sentiment. It is mainly used to analyze ratings and reviews. Typically this type of feedback is in different categories like the star rating system (1-5), where numbers indicate 1: very positive, 2: positive, 3: neutral, 4: negative, and 5: very negative.

### 3.2   Emotion detection

The name itself describes the function of this category and it helps to determine the emotion hidden behind the texts. The popular ones are anger, sadness, happiness, frustration, fear, panic, worry, or anxiety.

### 3.3   Aspect based

This sentiment analysis technique focuses more on the aspects of a particular product or service. To make it easier to understand, let us take an example of a LED television. The manufacturing company can ask for feedback on light, sound, picture quality, or durability and this will help the manufacturer/seller understand the issue with the product and improve it to make it better and more useful.

### 3.4   Intent analysis

By using this method, we can dig into a customer's intent. We can understand if the customer just wants information about the product or wants to purchase it. With the intent analysis, we can record, track, or form a pattern. This information can be used for target marketing.

### 3.5   Four basic metrics

In our current work, we have considered four basic sentiment metrics, each evaluated independently across different models, as follows. This particular choice was driven by the way sentiment analysis is structured by the model providing the best performance in the end, so the outputs of the other models were aligned to that.

**Sentiment.** overall or compound sentiment/polarity in range [-1.0,+1.0], so its value can be either negative or positive; some of the models can provide this metric and for other models it can be computed as a sum of the *positive* and the *negative* sentiment.

**Positive.** canonical positive sentiment assessment in range [0.0,+1.0], so its value can be only positive; some of the models can provide this metric and for other models it can be assessed as the *sentiment* if the value of the latter is above zero or zero otherwise.

**Negative**. canonical negative sentiment assessment in range [-1.0,0.0], so its value can be only negative; some of the models can provide this metric and for other models it can be assessed as the *sentiment* if the value of the latter is below zero or zero otherwise.

**Contradictive.** mutual constructiveness of the *positive* and *negative* assessments computed as *SQRT(positive * ABS(negative))*.

That is, instead of addressing the SA problem as a plain classification ('Positive' vs. 'Negative' vs. 'Neutral'), we have treated it as a multinomial classification problem in four independent dimensions corresponding to the individual metrics mentioned above.

## 4 Model Evaluation Experiments

We ran the same data through 21 different sentiment models for our experiments, calculated the sentiment score, and compared them. The selected winning models have been fine-tuned and re-evaluated, so overall 22 individual models are presented in Figure 1. All of the evaluated models are publicly available following the respective references.

### 4.1 Data

We have used about 100,000 news items (tweets and Reddit posts) across 77 public Twitter timelines and Reddit subreddits over the six month period of July to December of 2021 for exploration of the connection between the sentiment and the price movement discussed at the end of this paper. The data collection process has been based on official Reddit and Twitter APIs and was performed exclusively on public posts in public feeds. For the purpose of the algorithm quality assessment, we have used 490 tweets/posts from 5 randomly selected Twitter public feeds. The tweets/posts have been manually classified for both positive and negative sentiment in the range [-1.0,0.0] and [0.0,+1.0] respectively by two independent reviewers and made the "ground truth" sentiment assessment as the average of the two assessments for *positive* and *negative* metrics. Respectively, the "ground truth" for *sentiment* and *contradictive* metrics have been computed according to section 3.5. The list of source feeds as well as the reference corpus of manually classified feeds is available upon request.

### 4.2 Models

We tried our experiments evaluating the sentiment from raw textual data on a total of 21 different models.

**Afinn.** It was created by Finn Årup Nielsen; it is a lexicon-based approach and it has a total of 3,382 positive and negative words. Each word has a positive or negative score associated with it. The range for Afinn varies between -5 to 5 [1].

**Vader.** VADER stands for (Valence Aware Dictionary and sEntiment Reasoner). It was created by C.J. Hutto & E.E. Gilbert at the Georgia Institute of Technology. It is a lexicon and rule-based sentiment model specially created for texts in social media. It has over 9,000 words, and every word was marked by ten independent people from -4 (extremely negative) to 4 (extremely positive) and after that, the final score is the average of all 10 scores [2].

**TextBlob.** TextBlob is a lexicon and rule-based sentiment model. It has over 2,500 words and returns the subjectivity and polarity of the text. The polarity range lies between -1 (extremely negative) and 1 (extremely positive).

**GoogleNLP.** As the name indicates, GoogleNLP is owned by Google and it has a straightforward API to use. Google provides a free account for one month. Addition-ally, the model is a complete black box for the user, the sentiment score ranges from -1 to 1.

**AWS.** Amazon continuously increases its presence in machine learning and deep learning fields by providing various services. AWS comprehend is one of the services specifically for Natural Language Processing (NLP). It is also an utterly black-box model for the user, and Amazon provides a trial account for one month.

**Aigents.** Aigents is an "interpretable" model based on "n-grams," available as part of https://github.com/aigents/aigents-java distribution, and written in Java which comes with "out-of-the-box" vocabularies for n-grams associated with positive and negative sentiment. It has over 8,200 negative and over 3,800 positive n-grams and returns the overall sentiment/polarity of the text based on the frequencies of occurrences of the reference n-grams in the text along with independent positive and negative sentiment metrics. One of the specifics of the model is implementation of the "priority on order" principle as discussed in [3]. In the Aigents-specific implementation it means precedence given for n-grams with higher "n", so whenever any n-gram is matched, all matches of any other n-grams being parts of the former n-gram are disregarded. For instance, if tetragram ["not","a","bad","thing"] is matched, then both bigram ["bad","thing"] and unigram ["bad"] are disregarded and discounted. Similarly, matching bigram ["no", "good"] disregards and discounts both constituent unigrams ["no'] and ["good"]. In addition to that, the model has an option to provide logarithmic scaling of the counted frequencies and our studies have revealed that by enabling this option it provides better performance.

**BERT based models.** In our experiments, we used 15 BERT-based models trained on different datasets.

*Distilbert-base-uncased.* It is a distilled version of BERT, it is 40% smaller and 60% faster than BERT and keeps 97% of BERT's language understanding. distilbert was trained on the same dataset as BERT, English Wikipedia and Toronto Book Corpus [4].

*finiteautomata/bertweet-base-sentiment-analysis.* a transformer-based library. The base model is BerTweet, a RoBERTa model. The model was trained on SemEval 2017 corpus (around ~40k tweets) [5].

*cardiffnlp/twitter-roberta-base-sentiment.* The base model used was RoBERTa and was trained on ~58 Million tweets [6].

*ProsusAI/finBERT.* The base model used was BERT, and it was created to analyze financial texts. It was trained using the TRC2-financial dataset, which has 400K sentences, Financial PhraseBank which has 4845 sentences from financial news and FiQA Sentiment dataset [7].

*moussaKam/barthez-sentiment-classification.* The base model was BERT, it is a seq2se2 model for french [8].

*textattack/bert-base-uncased-imdb.* The researchers created a python framework "TextAttack" which is used for adversarial training, data augmentation in NLP the base model is BERT and trained on IMDB dataset [9].

*initeautomata/beto-sentiment-analysis.* Transformer-based library used for sentiment analysis, emotion analysis, and hate speech detection, trained on TASS 2020 corpus (around ~5k tweets) [14].

*siebert/sentiment-roberta-large-english.* This model is a finetune of RoBERTa-large [10] and it was trained and evaluated on 15 diverse datasets [11].

*sagorsarker/codeswitch-spaeng-sentiment-analysis-lince.* A BERT-based model used for language identification, pos tagging, name entity recognition, sentiment analysis. It was trained on LinCE[13] dataset and can be used on mixed languages: English, Spanish, Hindi and Nepali.

*aychang/roberta-base-imdb.* roBERTa was used as a base model and trained on the IMDB dataset.

*rohanrajpal/bert-base-multilingual-codemixed-cased-sentiment.* base model used was bert-base-multilingual-cased and fimetuned on SAIL 2017 dataset [12].

*abhishek/autonlp-imdb_sentiment_classification-31154.* BERT based model trained on IMDB dataset.

*VictorSanh/roberta-base-finetuned-yelp-polarity.* This model is based on RoBERTa and fine tuned on the Yelp polarity dataset.

*severo/autonlp-sentiment_detection-1781580.* BERT based model trained on IMDB dataset. Model accuracy 0.9426 and Precision: 0.930.

*mrm8488/distilroberta-finetuned-tweets-hate-speech.* distilroberta-base fine-tuned on tweets_hate_speech_detection dataset for hate speech detection.

## 5 Roadblocks

During these experiments, we encountered a great deal of challenges.

**Sarcasm.** People use sarcasm in their posts or conversation, it is the way of expressing a negative sentiment using a backhanded compliment. This situation can make it difficult for the sentiment model to understand the true context of the texts. If most of the texts contain sarcasm, it results in a higher number of positive sentiments even though in reality, it was negative.

**Idioms.** Sentiment analysis methods are still not mature enough to understand the idioms used in the texts.

**Negations.** High use of negation leads to misclassification. For example "not bad" is positive but for most of the lexicon-based models it will be negative because we are using a negative word with negation. This word order makes it positive, but most lexicon-based models consider it negative.

**Non-text data.** Twitter and Reddit are not limited to texts only. Users can upload audio, images, and videos. If the images contain a strong indication of price change, the sentiment model will miss that.

## 6 Experimental Results

The evaluation of the 21 models has been performed relying on the "ground truth" reference data discussed in section 4.1 with results presented in Figure 1.

The winning ("aigents" in Figure 1) model has also been used for fine-tuning, so "out-of-the-box" vocabularies were updated to get in sync with cryptocurrency do-main terminology and jargon. This has become possible due to the "interpretable" nature of the Aigents model. For the purpose of the fine-tuning, the results of the sentiment analysis, referencing 490 tweets were spotted for the misalignments between "predicted" values of *positive* and *negative* metrics and their respective "ground truth" counterparts with the discrepancy exceeding *0.5* for any of the two metrics. Furthermore, the content of the corresponding texts were considered as a clue to search for subject domain area terminology, jargon and figures of speech to add respective n-grams to either positive or negative vocabulary. Finally, the updated vocabularies were used to re-evaluate the model ("aigents+" in Figure 1) so we have received 22 individual models in the end. The latter fine-tined model is available as open source.

In addition to that, we have tried to build "ensemble" models, using all 22 models and only the top 3 models selected based on their superior performance, seen as "ensemble(all)" "ensemble(all)" "ensemble(top 3)" in Figure 1, respectively.

The performance of the models has been evaluated using the Pearson correlation coefficient across 490 reference tweets/posts for each of the four metrics between the values "predicted" by the model and the "ground truth" values. The average correlation over the four metrics was used as a score across all models as presented in Figure 1. As we can see in Figure 1, the top performance according to the Pearson correlation corresponds to fine-tuned Aigents model (0.57). Next, the "out-of-the-box" Aigents model (0.33) lines up with the finBERT model pre-trained on the financial domain (0.32). The other remaining models either barely approach the threshold of 0.3 or stay behind around a level of 0.0 showing no correspondence to the "ground truth" assessments.

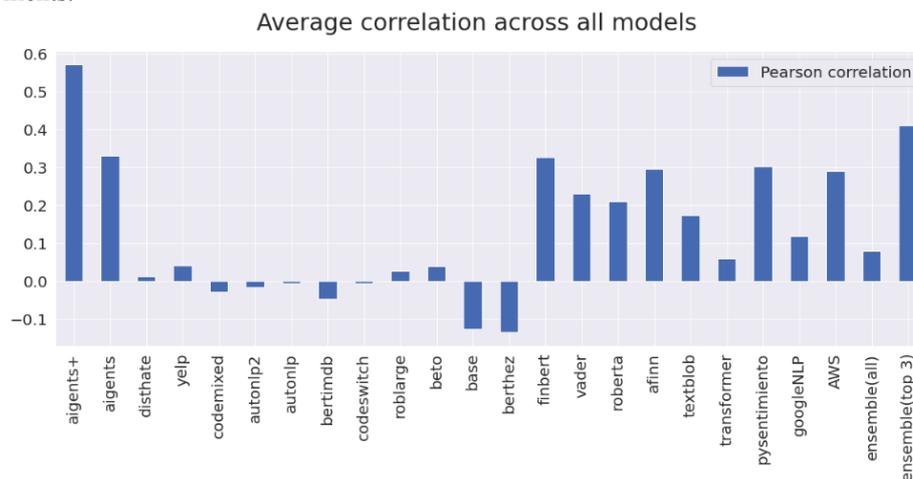

**Fig. 1.** The bar chart above shows the average Pearson correlation between sentiment metrics "predicted" by respective models and "ground truth" provided by humans. We can see the "out-of-the-box" Aigents model "aigents" has a correlation of ~0.33, and after fine-tuning, "aigents+" has a correlation of ~0.57. "ensemble(all)" corresponds to average metrics across all models, and "'ensemble(top 3)" corresponds to the average of the best three models (aigents+, aigents and finBERT).

Moreover, the whole volume of data of 100,000 tweets and Reddit posts across 77 public Twitter timelines and Reddit subreddits over the six month period has been used to search for a connection between the sentiment metrics and the price moves, following the concepts of causal analysis on time series discussed in [15], as shown in Figure 2. We can see that the plots corresponding to overall sentiment and positive metrics are presenting the peaks in the correlation value at -2 days shift. Moreover, the plot for the contradictive metric is showing the peak at -2 and -1 days shift while for the negative metric we see the high negative correlation at -1 day. Although the correlation values are not high (about 0.15), the amount of underlying data volume may suggest potential causal connections between respective sentiment metrics and the price change with two or one-day lag.

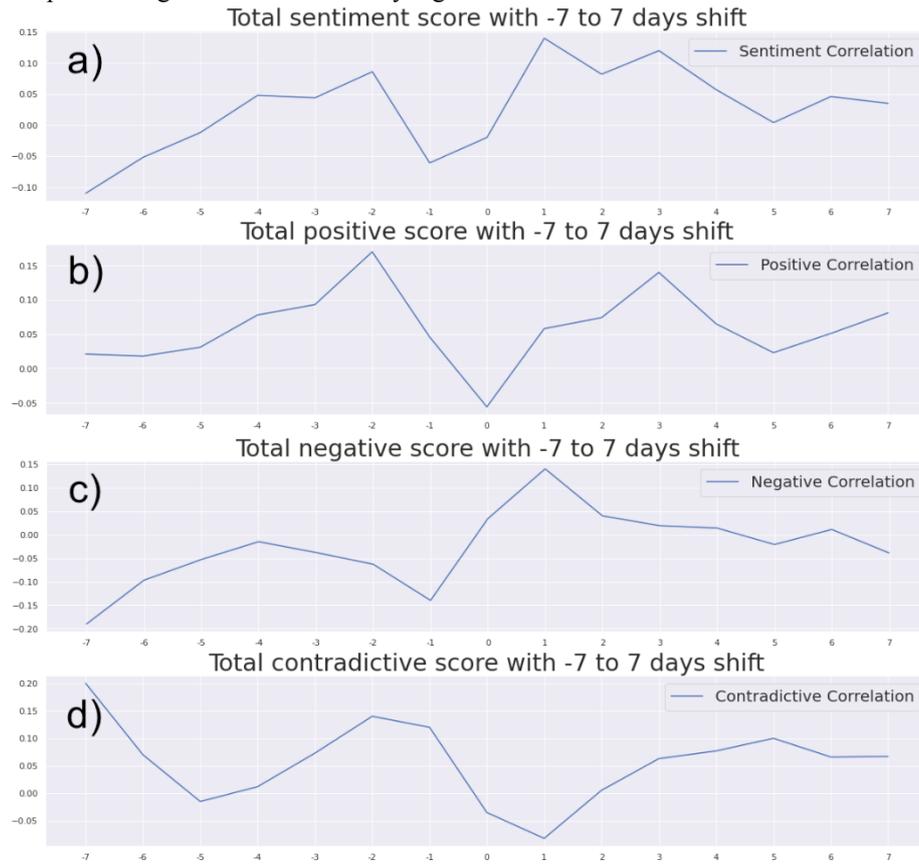

**Fig. 2.** Temporal correlation analysis for different sentiment metrics with mutual Pearson correlation computed between the daily Bitcoin price difference (derivative) and respective metrics over six months from July to December 2021, computed using the "aigents+" model with relative lags (shifts) of the price difference time series a certain number of days back or forward (-7 to +7) along the timeline, x-axis showing the days lag and y-axis corresponding Pearson correlation: a) Overall sentiment (positive + negative); b) Positive sentiment; c) Negative sentiment; d) Contraindicative (SQRT(positive * ABS(negative))).

In order to explore this possibility further, we have run the study following the concept of causal analysis in time series [15] across all four metrics evaluated. For each of the individual 77 news channels, we have explored 308 individual time series of sentiment metrics as potential causal sources of the single price difference time. We have run the temporal causation study evaluating different time lags in days [-10,+10] computing mutual Pearson correlation between each of the 308 potential causes and the price difference and retaining the weights of the computed value P(l,c,m) for every time lag l, news channel c, and metric m. Also, the channels c were weighted as W(c) according to the percentage of days with news present on such days. Then, for every lag l, the compound metric time series Y(l,d) = ΣX(c,m,d)*P(l,c,m)*W(c) for every day d have been built from the original raw metrics X(c,m,d). The compound metric building process was implemented starting from channels with the highest W(c) and P(l,c,m) adding ingredients up to Y(l,d) incrementally, as long as the correlation between the target price difference function and the current content of summed up Y(l,d) series for given time lag l keeps increasing. In the end, we have evaluated the terminal (maximum) correlation values for every lag, as shown in Figure 3.

Given a much clearer maximum at -1 day lag with correlation value as high as 0.55, compared with values corresponding to other lags, we can assume that selective inclusion and weighting of the news metrics and channels enables finding more causally connected time series, which builds up compound sentiment indicators that are potentially valuable for further feature engineering for the price prediction purposes. It can only suggest that the day before the cryptocurrency (Bitcoin) price change might be the most impactful from the perspective of the manipulative effect of social media on the market behavior.

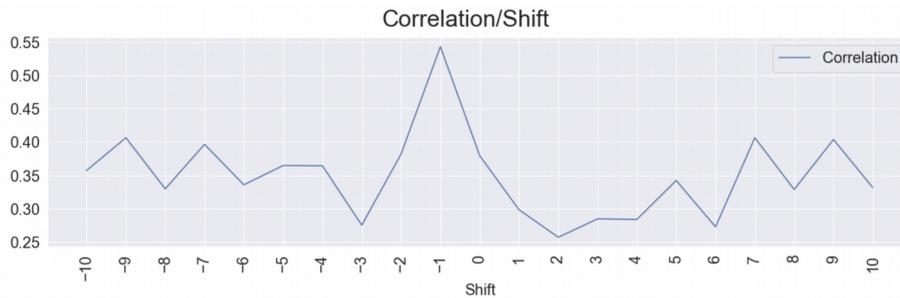

**Fig. 3.** Temporal correlation analysis for different sentiment metrics with mutual Pearson correlation computed between the daily Bitcoin price difference (derivative) and compound sentiment indicator built up upon 77 news channels and 4 sentiment metrics individually for respective time lags (in days, -10 to +10) over six months from July to December 2021, x-axis showing the days lag and y-axis corresponding Pearson correlation.

## 7     Conclusion and Future Work

In this paper, we have found the most reliable model for social media sentiment analysis in the cryptocurrency domain. We have shown how an "interpretable" sentiment analysis model could be significantly improved manually and without the huge costs for training the domain-specific corpus and creating and tagging this corpus for said purpose. In our further work, we are exploring how to automate this process of using the price movements being an implicit tagging of the sentiment-rich text data and learning the indicative n-grams from the temporally aligned market and news media data, with the option for manual review on the discovered patterns within the "inter-pretable" mode. We are looking forward to improving the performance of the best model further.

Additionally, we have preliminary explored the potential causal connection between social media sentiment and the price movements as an increase of expression of particular sentiment metrics two or one days before corresponding changes in price. We have also shown that the automated process of building compound sentiment indicators can be employed to increase the power of such connections. Our future work in this area will be dedicated to exploring the predictive power of the connection to improve the reliability of the price prediction and business applications for decentralized finance relying on such predictions.